\documentclass[runningheads]{llncs}
\usepackage{graphicx}
\usepackage{hyperref}
\usepackage{multirow}
\begin{document}
\title{More Than Meets the Eye: Analyzing Anesthesiologists’ Visual Attention in the Operating Room Using Deep Learning Models
}
\titlerunning{More Than Meets the Eye}
%
\makeatletter
\newcommand{\printfnsymbol}[1]{%
  \textsuperscript{\@fnsymbol{#1}}%
}
\makeatother

\author{Sapir Gershov\thanks{equal contribution} \inst{1}%
\and Fadi Mahameed\printfnsymbol{1} \inst{2,3}%
\and Aeyal Raz\inst{3}%
\and Shlomi Laufer\inst{2}%
}

\authorrunning{S. Gershov et al.}

\institute{Technion Autonomous Systems Program, Israel Institute of Technology, Technion City, Haifa 3200003, Israel \and
The Faculty of Data and Decision Sciences, Israel Institute of Technology, Technion City, Haifa 3200003, Israel \and
Rambam Health Care Campus, Haifa 3109601, Israel
}

\maketitle              
\begin{abstract}
Patient’s vital signs, which are displayed on monitors, make the anesthesiologist’s visual attention (VA) a key component in the safe management of patients under general anesthesia; moreover, the distribution of said VA and the ability to acquire specific cues throughout the anesthetic, may have a direct impact on patient’s outcome.
Currently, most studies employ wearable eye-tracking technologies to analyze anesthesiologists' visual patterns. Albeit being able to produce meticulous data, wearable devices are not a sustainable solution for large-scale or long-term use for data collection in the operating room (OR). Thus, by utilizing a novel eye-tracking method in the form of deep learning models that process monitor-mounted webcams, we collected continuous behavioral data and gained insight into the anesthesiologist’s VA distribution with minimal disturbance to their natural workflow.
In this study, we collected OR video recordings using the proposed framework and compared different visual behavioral patterns. We distinguished between baseline VA distribution during uneventful periods to patterns associated with active phases or during critical, unanticipated incidents.
In the future, such a platform may serve as a crucial component of context-aware assistive technologies in the OR.

\end{abstract}

\section{Introduction}
\subsection{Situation awareness and visual attention}
Situation awareness (SA) has been recognized as a crucial element in various domains, including aviation and sports, and has remained a subject of consistent interest in recent decades.\cite{Endsley2017TowardSystems,Stanton2017State-of-science:Systems}. Endsley defined the concept of SA as “the perception of elements in the environment within a volume of time and space, the comprehension of their meaning, and the projection of their status in the near future.”\cite{Endsley1988DesignEnhancement} It can be understood that SA is composed of three discrete hierarchies that are interdependent, with the first hierarchy-"perception", serving as the cornerstone of the structure. 
Throughout the following years, Endsley's work and research were adopted by various medical fields; anesthesiology pioneered its incorporation into the training process of residents.\cite{Gaba1995SituationAnesthesiology,Schulz2013SituationAnesthesia}
While under anesthesia, there is a multitude of clinical information pertaining to the patient that the anesthesiologist must monitor and oversee. Given that most of the information is presented visually, anesthesiologists' visual attention(VA) is essentially the method by which their perception is achieved, and on that rest, clinical decisions that affect patient care and safety; moreover, the way anesthesiologists distribute their VA both spatially and temporally in order to acquire specific cues along the provision of anesthesia may have a direct impact on the ability to provide better care.
From an observer's standpoint, such a phenomenon might even offer insight into the situation in which the provider is present and even divulge a part of his cognitive process.\cite{Gaba1995SituationAnesthesiology}

\subsection{Monitoring anesthesiologists visual behavior}
Anesthesiologists' VA has been the subject of many studies in the past. Schulz et al. \cite{Schulz2011VisualIncidents} used a head-mounted eye-tracking camera system to assess Anesthesiologists' distribution of VA to monitors, patients, and the environment. It was found that 20\% of visual attention was directed to the patient monitor during an uneventful procedure, with an increase to 30\% during critical incidents; they concluded that there is a correlation between visual attention and critical events requiring interventions.
White et al. \cite{White2018GettingResuscitations} and Roche et al. \cite{Roche2022AnesthesiaStudy} also examined the visual patterns of expert physicians; however, they used mobile eye-tracking glasses in simulated and actual resuscitation cases. Both studies reported similar results where expert physicians seemed to maintain situational awareness by using their ability to fixate on the monitor and vital signs temporarily, meanwhile, managing a specific task during the resuscitation. Furthermore, a study by Szulewski et al. \cite{Szulewski2019APerformance}, which also employed gaze-tracking glasses, reported evidence of visual patterns associated with better performance during simulated resuscitation scenarios.
Studies such as these lay the foundations for the claim that it is possible to utilize VA and SA for Anesthesiology residents' training and assess their advancement by analyzing their visual patterns. However, these studies assessed anesthesiologists' gaze using wearable eye-tracking devices. Although these devices produce highly accurate data, they have an inherent flaw as they are not a sustainable and ecological solution for long-term data collection in an environment such as the operating room (OR). Due to the devices' limited battery life, using them during lengthy procedures may be challenging. Moreover, based on what was reported, they require calibration before use, which typically requires additional staff and might impede workflow in the OR. In addition, these devices are inconvenient for the anesthesiologists who need to wear them during their long and intensive working hours \cite{Wagner2022Video-basedStudy}.
Our study presents an alternative approach that utilizes advanced Deep Learning techniques that provide continuous, day-to-day visual behavioral data with minimal interference in the daily workflow. If successful, such a system will facilitate the collection of vast amounts of data, enable in-depth analysis of the anesthesiologists' work, and potentially lay the cornerstone for developing context-aware assistive technologies in the OR. Implementing a webcam-based eye contact recognition model suitable for medical simulations and real-life OR settings. The presented framework will facilitate empirical comparisons between visual behaviors and patterns, whether uneventful or active for one individual during the procedure or between two individuals' patterns in a similar scenario (e.g., a resident and his attending). In the future, this may help determine the VA effect on patient care.   

\subsection{Challenges in gaze pattern detection}
There are two primary inquiries regarding the gaze of Anesthesiologists: whether they have inspected the patient's monitor and at what point this examination took place. This differentiation will determine if the physician noticed the critical cue displayed on the patient's monitor. We addressed these questions using the "Onfocus" detection, which identifies whether the individual's focus is on the camera \cite{Zhang2022OnfocusImages}.
Onfocus detection in unconstrained capture conditions presents numerous challenges arising from complex image scenes, inevitable occlusions, varied face orientations, continuous changes in frame focus, multiple objects appearing, and factors such as blur and over-exposure. Zhang et al. \cite{Zhang2022OnfocusImages} introduced a model and dataset to assess onfocus detection under these challenges; however, certain components were lacking to make it suitable for clinical usage.
In the present study, we enhanced and adapted the model proposed by Zhang et al. to be suitable for implementation in the OR. Once we achieved a functional system, we applied our model to a newly generated dataset consisting of webcam videos capturing the gaze of anesthesiologists during medical simulations and real-life OR scenarios.
To summarize, the objectives of this study are twofold: to advance gaze detection methodology and enhance healthcare training and assessment via gaze pattern analysis. By employing unobtrusive equipment to detect gaze within the dynamic and demanding OR environment, and subsequently analyzing the context-specific VA patterns of anesthesiologists, this research presents an exciting opportunity to enhance deep learning models for detecting focus. Furthermore, it offers valuable insights into the intricate work patterns involved in anesthesia delivery without compromising patient safety and potentially even improving it.

\section{Material}
\subsection{Onfocus Detection In the Wild dataset}
Zhang et al. \cite{Zhang2022OnfocusImages} presented a significant contribution to onfocus detection by introducing the large-scale 'OFDIW' dataset. The OFDIW dataset has unique videos featuring face-to-camera communication, a single camera perspective, fully visible faces, and minimal frame-to-frame variations. However, the dataset's limitations, including using a single camera with fixed camera-face orientation, result in the absence of crucial characteristics necessary for comprehensive VA analysis in the OR.


\subsection{Eye-Context Interaction Inferring Network}
Zhang et al.'s \cite{Zhang2022OnfocusImages} work provided a novel end-to-end model for onfocus detection. The model, named "Eye-Context Interaction Inferring Network" (ECIIN), is a deep learning architecture composed of a convolution neural network (CNN) context capsule network (CAP) \cite{Ramasinghe2019AClassification}. Given an input frame, the ECIIN model begins by localizing the eye regions, which are then processed with conventional convolution layers adopted to extract feature maps (Eye CNN). In parallel, context features from the original image are extracted using another CNN (Context CNN). The feature vectors are passed to the CAPs trained to reflect the context and eye status, respectively; furthermore, CAPs model the interaction between the eye and context regions. Eventually, the model classifies the eye-context interaction as "onfocus" or "out of focus". As it shows, Zhang et al.'s model does not take advantage of Yang et al.'s publicly available dataset for face detection \cite{Yang2016WIDERBenchmark}, which is rich with labeled data suitable for training a model for such a task. They also do not employ a well-known, state-of-the-art detection model that can be more accurate and robust using transfer learning techniques. Finally, their model does not support detecting multiple objects in the same image. 

\subsection{WIDER FACE dataset}
The WIDER FACE dataset\cite{Yang2016WIDERBenchmark} is one of the most extensive publicly available datasets for face detection. It contains 32,203 images and 393,703 labels of faces with a wide range of scale, poses, and occlusion variability.
Each recognizable face in the WIDER FACE dataset is labeled by bounding boxes, which must tightly contain facial landmarks (FL) (e.g., forehead, chin, and cheek). In the case of occlusion, the face is labeled with an estimated bounding box.
\subsection{YOLOv7 for Face Detection}
YOLO, which stands for “You Only Look Once”, is a popular family of real-time object detection algorithms. The original YOLO object detector was published by Redmon et al. \cite{Redmon2016YouDetection}. Since then, different versions and variants of YOLO have been proposed, each providing a significant increase in performance and efficiency.
Previously, Qi et al. \cite{Qi2021YOLO5Face:Detector} published a modification of the YOLO architecture, YOLO5Face, which treats face detection as a general object detection task. In their work, they designed a face detector model capable of achieving state-of-the-art performance in varying image sizes by adding a five-point landmark regression head into the original architecture and using the Wing loss function \cite{Feng2018WingNetworks}. 



FL detection can be achieved by Dlib-ml \cite{King2009Dlib-ml:Toolkit}, a cross-platform, open-source software library with pre-trained detectors for FL. The Dlib detector estimates the location of 68 coordinates $(x, y)$ that map the facial points on a person’s face. Though newer algorithms leverage a dense “face mesh” with machine learning to infer the 3D facial surface from single camera input, these models fail to produce superior results when the acquired images have disturbances and motion \cite{Deng2020Retinaface:Wild}.

\subsection{Spatiotemporal Gaze Architecture}
Most works that tackled the task of detecting gaze target prediction constructed 2D representations of the gaze direction, which fails to encode whether the person of interest is looking onward, backward, or sideward. Chong et al. \cite{Chong2020DetectingVideo} proposed using a deep-learning network to construct a 3D gaze representation and incorporate it as an additional feature channel. The input to their network was the video frame scene, the heads' position in the frame, and the reciprocal cropped head images; however, they did not provide a face-detection model to generate this input automatically. In addition, Chong et al.'s work applied $\alpha$ - a learned scalar that evaluates whether the person’s object of attention is inside or outside the frame, with higher values indicating in-frame attention. Yet, when the person’s object of attention was outside the frame, they did not examine the cases in which the object of attention was the camera itself.

\section{Methods}
\subsection{Pipeline construction}
\begin{figure}[ht!]
    \centering
    \includegraphics[scale=0.36]{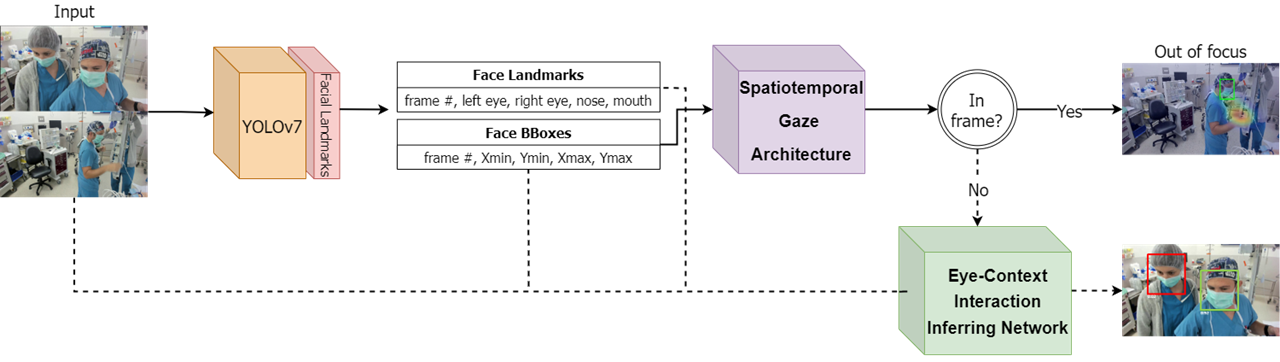}
    \caption{End-to-end framework pipeline. The Spatiotemporal prediction is indicated by a bounding box over the allocated head and a heatmap over the object of attention. If the attention is on an object inside the frame, we automatically classify the image as "out of focus". Otherwise, we pass the frame to the ECIIN model. 
    The ECIIN network classification confidence is indicated by color, where green is for high confidence and red is for low confidence. The score next to the bounding box is the prediction probability.} 
    \label{fig:1}
\end{figure}

For face detection, we trained YOLOv7 on the WIDER FACE dataset \cite{Yang2016WIDERBenchmark}. For each detected bounding box prediction, an FL algorithm was applied \cite{King2009Dlib-ml:Toolkit}. We used only the coordinates visible in the collected data - eyes, nose, and mouth. 
The coupling of YOLOv7 trained for face detection with an FL detector is a suitable replacement for Zhang et al.'s. \cite{Zhang2022OnfocusImages} ECIIN-designed network modules. Our modifications harness the benefits of well-trained object detectors and large datasets. Thus, produce superior results. Once the region of interest is located, we apply the process described in Zhang et al.'s. work to generate the onfocus detection.
Lastly, we modified Chong et al. spatiotemporal model \cite{Chong2020DetectingVideo} by introducing the ECIIN component. This modification has improved Chong et al. model performance in cases where the object of attention is "out-of-frame". 
The complete end-to-end pipeline is depicted in Figure \ref{fig:1}. 

Applying the end-to-end pipeline to video footage produced a new video with labeled bounding boxes around faces, indicating whether they were classified as "onfocus". A text file was also generated, documenting the model predictions for each frame. Since multiple "onfocus" labels with varying confidence levels could appear in each frame, we established a threshold of $0.72$ to exclude predictions with low confidence. This value was achieved via hyperparameter tuning of the framework. Addressing the output text file as a time series enabled various statistical measurements to be applied. 

\subsection{Data acquisition}
\subsubsection{Simulation data}
\begin{figure}[]
    \centering
    \includegraphics[scale=0.35]{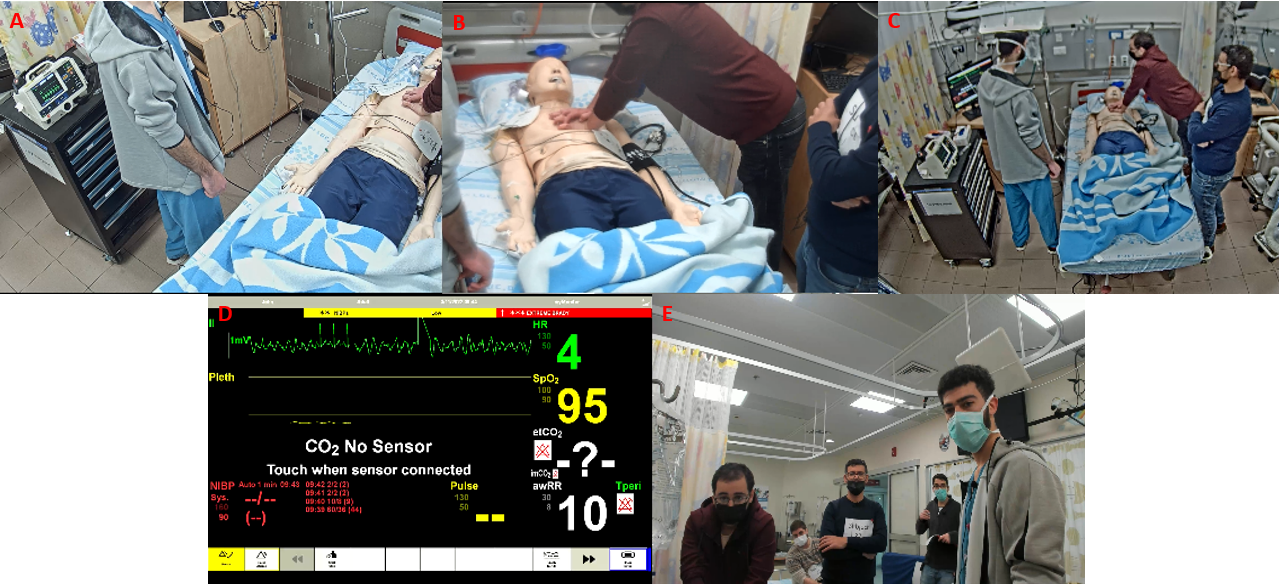}
    \caption{Data acquisition system.(A) Nurse working area; (B) Physician working area; (C) Overview of the simulation area; (D) Patient monitor; (E) Patient monitor point-of-view.
    The participants gave their approval for publishing these images.}
    \label{fig:2}
\end{figure}

In order to adjust the proposed framework to the OR settings, we began with a small-scale dataset of residents' VA during simulated medical scenarios. A webcam was attached to a patient monitor, which recorded the anesthesiologists' VA patterns. Additional vantage points were added in order to offer a broader clinical context (see Figure \ref{fig:2}).

\begin{figure}[h]
    \centering
    \includegraphics[scale=0.35]{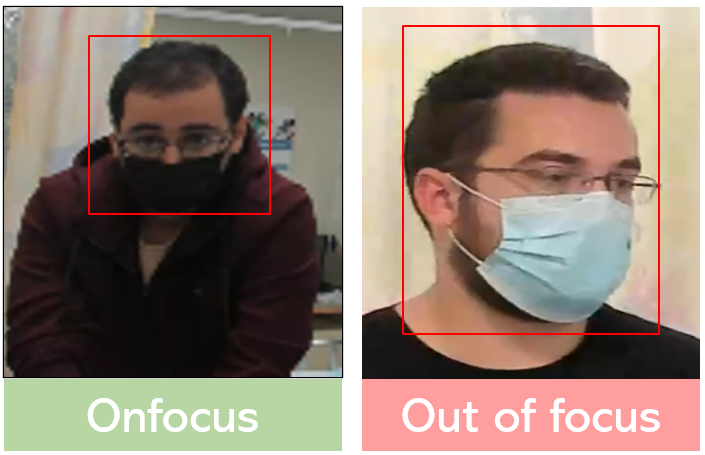}
    \caption{Zoom-in examples of the labeled frames.}
   \label{fig:3}
\end{figure}
The Video and audio data collecting was managed via StreamPix digital video recording software developed by NorPix Inc..
The dataset included 20-minute video recordings of 31 simulations performed by 33 residents, from which over 1200 frames were extracted. The appearing faces were manually labeled with bounding boxes suitable for the YOLO network. An independent human observer marked the events of direct eye contact with the monitor; these events were classified as "Onfocus" or "Out of focus" while maintaining a balanced and diverse dataset (see Figure \ref{fig:3}). The labeled frames were used for fine-tuning the proposed framework, which was then applied to video recordings from the OR.

\subsubsection{OR data}
\begin{figure}[h]
    \centering
    \includegraphics[scale=0.4]{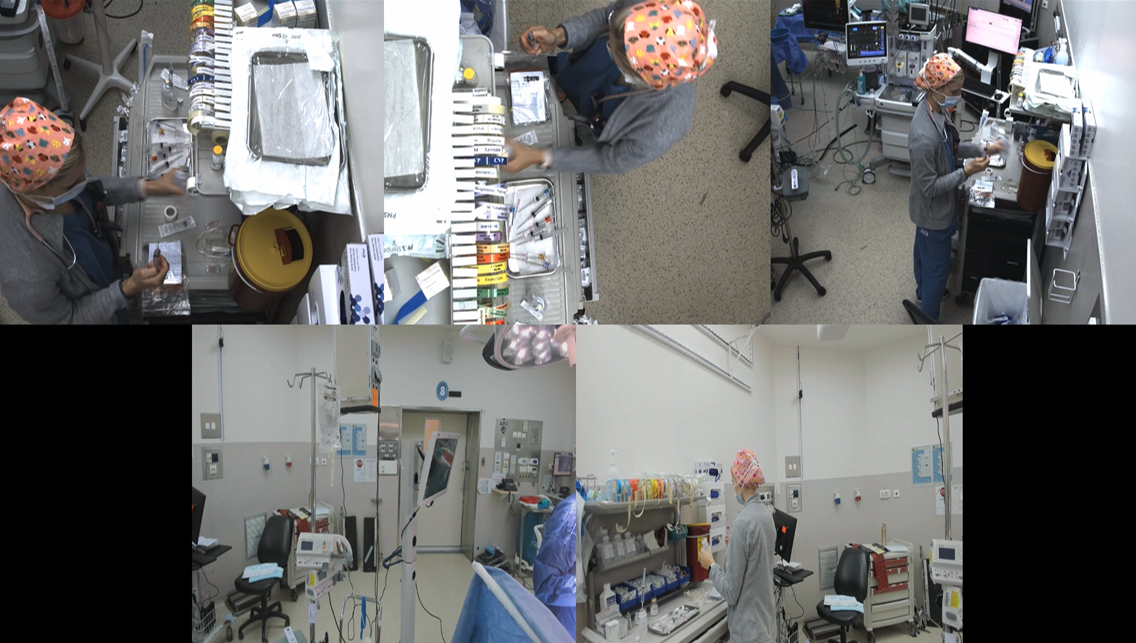}
    \caption{main angles of the OR RGB camera system set-up}
   \label{fig:4}
\end{figure}

After analyzing the preliminary data and acquiring the approval of the Rambam Health Care Campus Institutional Review Board (IRB), we installed a similar, larger-scale RGB camera system in the institution's OR. The system included more vantage points, see Figure \ref{fig:4}, such as the anesthesia cart (containing most medications and equipment), additional monitors (anesthesia machine/ventilator monitors), and different workstations. The collected data was analyzed and labeled by the proposed framework. 

A separate labeling of the OR dataset was performed; different tasks, behaviors, and monitor interactions of anesthesiologists were manually labeled by an independent researcher(also an anesthesiologist) using the event logging software for video/audio coding and live observations BORIS. The labeled data was then cross-referenced between those labeled manually and those labeled using the deep learning model.
Using this combination, we validated the applicability of the proposed framework in a real-life clinical environment.

\section{Results}
\subsection{Evaluation of the proposed framework}     
We evaluated the different models' performance on our labeled simulation frames. The dataset was randomly divided into 80\% train, 10\% validation, and 10\% test, followed by 5-fold cross-validation.
We used our YOLOv7 model for face detection to provide the input for the Spatiotemporal model \cite{Chong2020DetectingVideo} and then assessed the model's heatmaps. Theoretically, the Spatiotemporal model will not generate heatmaps in cases where the anesthesiologist has direct eye contact with the monitor. Thus, we can address the calculated heatmaps as binary classifications (i.e., the object of attention is inside the frame or outside).    
Table \ref{tab:1} shows the performance of the different models regarding the task of onfocus detection.

\begin{table}[h]
\caption{Models onfocus detection results on medical simulations frames.}
\label{tab:1}
\centering
\begin{tabular}{|c|c|c|c|}
\hline
\textbf{Model} & \textbf{Dataset} & \textbf{Accuracy} & \textbf{F1-Score} \\ \hline\hline
ECIIN             &  \multirow{3}{*}{Medical simulations} & $63.98\%$ $\pm 2.53\%$        & $0.64$ $\pm 0.03$         \\ \cline{1-1} \cline{3-4} 
Spatiotemporal    &                    & $71.03\%$  $\pm 1.87\%$       & $0.72$ $\pm 0.11$         \\ \cline{1-1} \cline{3-4} 
Complete pipeline &                    & \textbf{89.22\% $\pm$ 1.26\%} & \textbf{0.87 $\pm$ 0.02} \\ \hline
\end{tabular}
\end{table}

\subsection{VA analysis in the OR}
\begin{figure}[h]
    \centering
    \includegraphics[width=\textwidth]{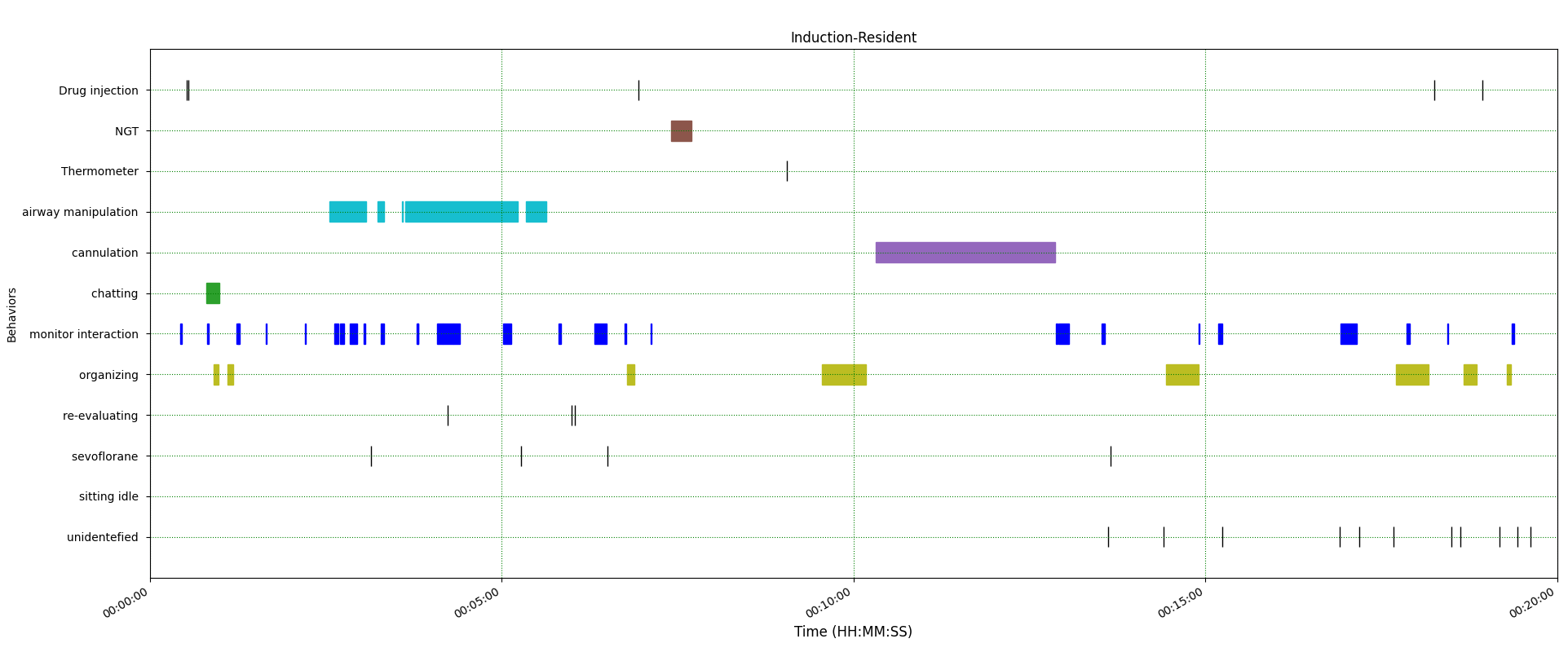}
    \caption{Human evaluator mapping of tasks distribution and duration along the time course of an uneventful induction. The provider is an anesthesia attending.}
    \label{fig:5}
\end{figure}

We conducted a detailed analysis of 12 anesthesia procedures, of which ten were conducted by residents, one by an attending doctor, and one by an attending doctor in collaboration with a resident. We applied the proposed framework to assess gaze frequency, length, and cumulative time during the induction phase of anesthesia. This phase typically commences with the acquisition of baseline vital sign measurements, drug administration to achieve sedation, management and securing of the patient's airway, and patient positioning in preparation for surgery.
Additional camera angles were used to categorize almost all activities related to the anesthesia induction phase, including interactions with the monitor, among other tasks. Detection of monitor interactions using each approach is compared in Table \ref{tab:2}. Figure \ref{fig:5} outlines the different tasks labeled by a human evaluator during a resident-conducted induction of a procedure.

\begin{table}[h]
    \centering
    \caption{Comparative results of the anesthesiologist VA - displayed as mean$\pm$SD, the total time displayed as the percentage of time throughout the induction phase.}
    \label{tab:2}
    \begin{tabular}{|c|c|c|c|c|c|}
    \hline  
    \textbf{Monitor} & \textbf{Detector} & \textbf{Freq. [$(5 min)^{-1}$]} & \textbf{Duration [s]} & \textbf{Total time (\%)} & \textbf{P-value} \\ \hline\hline
        \multirow{2}{*}{Patient}&Framework &14$\pm$3.78&4.59$\pm$1.23& 27.23\%$\pm$3.14\% &\multirow{2}{*}{\textbf{0.0167}}\\ \cline{2-5}
        &Human observer &11$\pm$5.96&4.33$\pm$2.14&23.81\%$\pm$5.41\%&\\ \hline
        \multirow{2}{*}{Ventilator}&Framework &24$\pm$3.93&6.13$\pm$1.24&34.59\%$\pm$11.83\% &\multirow{2}{*}{\textbf{0.271}}\\ \cline{2-5}
        &Human observer &23$\pm$6.89&10.19$\pm$3.11&47.39\%$\pm$4.70\%&\\ \hline
    \end{tabular}
\end{table}

\begin{table}[h]
    \centering
    \caption{Comparative results of the anesthesiologist VA - displayed as the percentage of time spent observing monitors throughout the airway manipulation tasks.}
    \label{tab:3}
    \begin{tabular}{|c|c|c|c|}
    \hline  
    \textbf{Monitor} & \textbf{Detector} & \textbf{Total time (\%)} & \textbf{P-value} \\ \hline\hline
        \multirow{2}{*}{Patient}&Framework & 38.72\%$\pm$9.04\% &\multirow{2}{*}{\textbf{0.0238}}\\ \cline{2-3}
        &Human observer &40.19\%$\pm$3.65\%&\\ \hline
        \multirow{2}{*}{Ventilator}&Framework &49.22\%$\pm$8.54\% &\multirow{2}{*}{\textbf{0.0946}}\\ \cline{2-3}
        &Human observer &53.07\%$\pm$6.11\%&\\ \hline
    \end{tabular}
\end{table}

Manual labeling served a dual purpose. On the one hand, it offered a "ground truth" for the data labeled by the proposed framework, specifically concerning monitor interactions. When comparing the monitor interaction data from both analytical approaches, the results were reasonably consistent regarding time spent interacting with the patient's monitor, as indicated in Table \ref{tab:2}. However, frequency, duration, and total time results did not align well for ventilator monitor interactions. This discrepancy can be partly attributed to the camera angle.
On the other hand, overlapping manually labeled events with monitor interactions detected by the proposed framework provided the procedural context for these interactions. For example, during the Airway manipulation task, which includes multiple steps such as Mask ventilation, endotracheal tube or laryngeal mask placement, and verification of adequate placement, significant findings were noted: the overall time dedicated to patient monitor interaction was considerably higher compared to other segments of the induction phase, as demonstrated in Table \ref{tab:3}.

\section{Discussion}
Considering the significance of visual attention (VA) in healthcare quality and patient safety, there is a growing belief that eye-tracking technology can provide valuable insights in these areas. As technology has advanced, researchers have increasingly employed eye-tracking devices to study VA and examine the activities of healthcare professionals in their natural work settings. This is particularly relevant in the field of anesthesiology. However, many of these devices have been intrusive and disrupted the natural workflow.

To address this issue, we conducted a study utilizing webcams to record the eye contact of anesthesiologists with the operating room (OR) monitors. This unobtrusive approach allowed us to gather continuous visual behavioral data without interfering with the workflow of the anesthesiologists. Furthermore, we analyzed the context in which such eye contact occurred.

The findings of our study using the proposed framework have significant implications for clinical research by providing insights into the relationship between interventions, behaviors, and patient outcomes. Additionally, assessing and providing feedback on anesthesiologists' performance without compromising behavior or workflow could usher in a new era in clinical training and assessment.

We presented preliminary data analyses based on a small sample, focusing on VA during the induction phase of anesthesia procedures. Our findings align with previous reports by Schulz et al.\cite{Schulz2011VisualIncidents}, showing similar VA allocation patterns during increased activity phases (30\% of VA directed towards patient monitor). Moreover, the frequency and duration of VA exhibited comparable values to those reported in the work of Manser et al.\cite{manser2002analysing}. Thus, our proposed framework, combined with manual labeling of the different activities of an anesthesiologist in the OR environment, offers a more robust and non-intrusive method for analyzing OR data.

\section{Ethical consideration}
First, we assert that all procedures contributing to this work comply with the ethical standards of the national and institutional committees on human experimentation and with the Helsinki Declaration.

Second, participants presented in un-blurred images provided explicit permission for these images.

Third, the framework we describe only documents the medical personnel without disturbing the daily workflow or harming the patient. 

\section{Acknowledgment}
This study was funded by The Bernard M. Gordon Center for Systems Engineering at the Technion.

\bibliographystyle{splncs04}
\bibliography{refrences}

\end{document}